\documentclass[letterpaper, 10 pt, conference]{ieeeconf}  % Comment this line out if you need a4paper

\IEEEoverridecommandlockouts                              % This command is only needed if 
\usepackage{dblfloatfix} 
\usepackage[utf8]{inputenc} % allow utf-8 input
\usepackage[T1]{fontenc}    % use 8-bit T1 fonts
\usepackage{hyperref}       % hyperlinks
\usepackage{url}            % simple URL typesetting
\usepackage{booktabs}       % professional-quality tables
\usepackage{amsfonts}       % blackboard math symbols
\usepackage{nicefrac}       % compact symbols for 1/2, etc.
\usepackage{microtype}      % microtypography
\usepackage{amsmath}
\usepackage{subcaption}
\usepackage{wrapfig}
\usepackage[font=small,labelfont=bf]{caption}

\usepackage{graphics} % for pdf, bitmapped graphics files
\usepackage{epsfig} % for postscript graphics files
\usepackage{amssymb}  % assumes amsmath package
\usepackage{wrapfig}
\usepackage{authblk}
\title{Imitation from Observation: Learning to Imitate Behaviors from Raw Video via Context Translation}

\author{YuXuan Liu${^{\dagger*}}$\thanks{* These authors contributed equally to this work.}, Abhishek Gupta${^{\dagger*}}$, Pieter Abbeel${^\dagger}{^\ddagger}$, Sergey Levine${^\dagger}$\\
$^\dagger$ UC Berkeley, Department of Electrical Engineering and Computer Science\\
$^\ddagger$ OpenAI\\
\texttt{\{yuxuanliu,abhigupta,pabbeel,svlevine\}@berkeley.edu} \\
}

\begin{document}

\maketitle

\begin{abstract}
Imitation learning is an effective approach for autonomous systems to acquire control policies when an explicit reward function is unavailable, using supervision provided as demonstrations from an expert, typically a human operator. However, standard imitation learning methods assume that the agent receives examples of observation-action tuples that could be provided, for instance, to a supervised learning algorithm. This stands in contrast to how humans and animals imitate: we observe another person performing some behavior and then figure out which actions will realize that behavior, compensating for changes in viewpoint, surroundings, object positions and types, and other factors. We term this kind of imitation learning ``imitation-from-observation,'' and propose an imitation learning method based on video prediction with context translation and deep reinforcement learning. This lifts the assumption in imitation learning that the demonstration should consist of observations in the same environment configuration, and enables a variety of interesting applications, including learning robotic skills that involve tool use simply by observing videos of human tool use. Our experimental results show the effectiveness of our approach in learning a wide range of real-world robotic tasks modeled after common household chores from videos of a human demonstrator, including sweeping, ladling almonds, pushing objects as well as a number of tasks in simulation.
\end{abstract}
\section{Introduction}

Learning can enable autonomous agents, such as robots, to acquire complex behavioral skills that are suitable for a variety of unstructured environments. In order for autonomous agents to learn such skills, they must be supplied with a supervision signal that indicates the goal of the desired behavior. This supervision typically comes from one of two sources: a reward function in reinforcement learning that specifies which states and actions are desirable, or expert demonstrations in imitation learning that provide examples of successful behaviors. Both modalities have been combined with high-capacity models such as deep neural networks to enable learning of complex skills with raw sensory observations \cite{imitation_learning_martial_hebert_drew_bagnell_uav, mnih, levinefinn16JMLR, Pomerleau:1989:AAL:89851.89891}.
One major advantage of reinforcement learning is that the agent can acquire a skill through trial and error with only a high-level description of the goal provided through the reward function. However, reward functions can be difficult to specify by hand, particularly when the success of the task can only be determined from complex observations such as camera images \cite{PerceptReward}.

\begin{figure}[!t]
  \centering
  \includegraphics[width=0.49\textwidth]{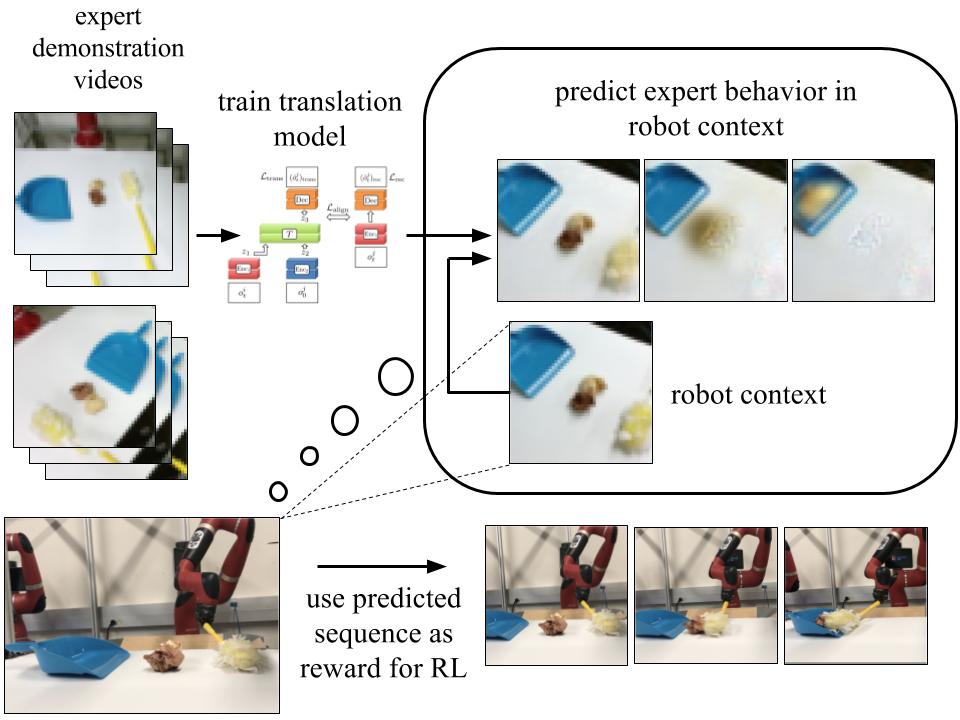} 
  \caption{Imitation from Observation using Context-Aware Translation. We collect a number of videos of expert demonstrations from a human demonstrator, and use them to train a context translation model. At learning time, the robot sees the context of the task it needs to perform. Then, the model predicts what an expert would do in the robot context. This predicted sequence is used to define a cost function for reinforcement learning thus enabling imitation from observation. The task shown here is illustrative of a wide range of tasks that we evaluate.}
  \label{fig:teaser}
\end{figure}

Imitation learning bypasses this issue by using examples of successful behavior. Popular approaches to imitation learning include direct imitation learning via behavioral cloning~\cite{ Pomerleau:1989:AAL:89851.89891} 
and reward function learning through inverse reinforcement learning~\cite{ngIRL}. Both settings typically assume that an agent receives examples that consist of sequences of observation-action tuples, and try to learn either a function that maps observations to actions from these example sequences or a reward function to explain this behavior while generalizing to new scenarios. However, this notion of imitation is quite different from the kind of imitation carried out by humans and animals: when we learn new skills from observing other people, we do not receive egocentric observations and ground truth actions. The observations are obtained from an alternate viewpoint and the actions are not known. Furthermore, humans are not only capable of learning from live observations of demonstrated behavior, but also from video recordings of behavior provided in settings considerably different than their own. Can we design imitation learning methods that can succeed in such situations? A solution to this problem would be of considerable practical value in robotics, since the resulting imitation learning algorithm could directly make use of natural videos of people performing the desired behaviors obtained, for instance, from the Internet.

We term this problem imitation-from-observation. The goal in imitation-from-observation is to learn policies only from a sequence of observations (which can be extremely high dimensional such as camera images) of the desired behavior, with each sequence obtained under differences in context. Differences in context might include changes in the environment, changes in the objects being manipulated, and changes in viewpoint, while observations might consist of sequences of images. We define this problem formally in Section~\ref{prelims}.

Our imitation-from-observation algorithm is based on learning a context translation model that can convert a demonstration from one context (e.g., a third person viewpoint and a human demonstrator) to another context (e.g., a first person viewpoint and a robot). By training a model to perform this conversion, we acquire a feature representation that is suitable for tracking demonstrated behavior. We then use deep reinforcement learning to optimize for the actions that optimally track the translated demonstration in the target context. As we illustrate in our experiments, this method is significantly more robust than prior approaches that learn invariant feature spaces~\cite{TPIL}, perform adversarial imitation learning~\cite{GAIL}, or directly track pre-trained visual features~\cite{sermanetRSS}. Our translation method is able to provide useful perceptual reward functions, and performs well on a number of simulated and real manipulation tasks, including tasks that require a robot to emulate human tool use. Videos can be found on \url{https://sites.google.com/site/imitationfromobservation/}
\section{Related Work}\label{sec:related}
Imitation learning is usually thought of as the problem of learning an expert policy that generalizes to unseen states, given a number of expert state-action demonstration trajectories~\cite{SchaalImitation,Argall}. Imitation learning has enabled the successful performance of tasks in a number of complex domains such as helicopter flight through apprenticeship learning ~\cite{AbbeelHeli}, learning how to put a ball in a cup and playing table tennis ~\cite{ballincup}, performing human-like reaching motions ~\cite{billardreaching} among others. These methods have been very effective however typically require demonstrations provided through teleoperation or kinesthetic teaching, unlike our work which aims to learn from observed videos of other agents performing the task. Looking at the imitation learning literature from a more methodological standpoint, imitation learning algorithms can largely be divided into two classes of approaches: behavioral cloning and inverse reinforcement learning.

Behavioral cloning casts the problem of imitation learning as supervised learning, where the policy is learned from state(or observation)-action tuples provided by the expert. In imitation-from-observation, the expert does not provide actions, and only provides observations of the state in a different context, so direct behavioral cloning cannot be used. Inverse reinforcement learning (IRL) methods instead learn a reward function from the expert demonstrations~\cite{ngIRL, AbbeelIRL, lpk2011, ZiebartMBD08}. 
This reward function can then be used to recover a policy by running standard reinforcement learning ~\cite{maxmarginIRL, ramchandranamir}, though some more recent IRL methods alternate between steps of forward and inverse RL~\cite{GCL, PI2IRL, REIRL, GAIL}. 
While IRL methods can in principle learn from observations, in practice using them directly on high-dimensional observations such as images has proven difficult. 

Aside from handling high-dimensional observations such as raw images, our method is also designed to handle differences in context. Context refers to changes in the observation function between different demonstrations and between the demonstrations and the learner. These may include changes in viewpoint, object positions, surroundings, etc. Along similar lines, \cite{TPIL} directly address learning under domain shift. However, this method has a number of restrictive requirements, including access to expert and non-expert policies, directly optimizing for invariance between only two contexts (whereas in practice demonstrations may come from several different contexts), and performs poorly on the more complex manipulation tasks that we consider, as illustrated in Section~\ref{sec:experiments}. \cite{sermanetRSS} proposes to address differences in context by using pretrained visual features, but does not provide for any mechanism for context translation, as we do in our work, relying instead on the inherent invariance of visual features for learning. Follow-up work proposes to further increase the invariance of the visual features through multi-viewpoint training~\cite{tcn}. \cite{handfuture} propose to learn robotic skills from first person videos of humans by using explicit hand detection and a carefully engineered vision pipeline. In contrast, our approach is trained end-to-end, and does not require any prior visual features, detectors, or vision systems. \cite{oneshotimitation} proposed to use demonstrations as input to policies by training on paired examples of state sequences, however our method operates on raw observations and does not require any actions in the demonstrations, while this prior method \
operates only on low-dimensional state variables and does not deal with context shift like our method.

Our technical approach is related to work in visual domain adaptation and image translation. Several works have proposed pixel level domain adaptation~\cite{cycleGAN, googpixelDA, pix2pix}, as well as translation of visual style between domains~\cite{gatys}, by using generative adversarial networks (GANs). The applications of these methods have been in computer vision, rather than robotic control. Our focus is instead on translating demonstrations from one context to another, conditioned on the first observation in the target context, so as to enable an agent to physically perform the task. Although we do not use GANs, these prior methods are complementary to ours, and incorporating a GAN loss could improve the performance of our method further.

In our work we consider tasks like sweeping, pushing, ladling(similar to pouring) and striking. Several prior methods have looked at performing tasks like these although typically with significantly different methods. Tasks involving cleaning with a brush, similar to our sweeping tasks was studied in ~\cite{XuC14} but is done using a low cost tool attachment and kinesthetic programming by demonstration. Besides \cite{sermanetRSS}, tasks involving pouring were also studied in ~\cite{Pouring} using a simple PID controller with a specified objective volume rather than inferring the objective from demonstrations. Similar flavors of tasks were also considered in ~\cite{deepMPC, robobarista}, but we leave those specific tasks to future work. Other work ~\cite{SE3} also considers tasks of pushing objects on a table-top but uses predictive models on point-cloud data and uses a significantly different intuitive physics model with depth data. 

\section{Problem Formulation and Overview}
\label{prelims}

In the imitation-from-observation setting that we consider in this work, an agent observes demonstrations of a task in a variety of \emph{contexts}, and must then execute the demonstrated behavior in its own context. We use the term context to refer to properties of the environment and agent that can vary across demonstrations, which may include the viewpoint, the background, the positions and identities of objects in the environment, and so forth. The demonstrations $\{D_1, D_2,....D_n\} = \{[o_0^1, o_1^1,.....o_T^1], [o_0^2, o_1^2,.....o_T^2],...., [o_0^n, o_1^n,.....o_T^n]\}$ consist of observations $o_t$ that are produced by a partially observed Markov process governed by an observation distribution $p(o_t | s_t, \omega)$, dynamics $p(s_{t+1} | s_t, a_t, \omega)$, and the expert's policy $p(a_t | s_t, \omega)$, with each demonstration being produced in a different context $\omega$. Here, $s_t$ represents the unknown Markovian state, $a_t$ represents the action (which is not observed in the demonstrations), and $\omega$ represents the context. We assume that $\omega$ is sampled independently from $p(\omega)$ for each demonstration, and that the imitation learner has some fixed $\omega_l$ from the same distribution. Throughout the technical section, we use $o_t^i$ to refer to the observation at time $t$ from a context $\omega_i$.

While a practical real-world imitation-from-observation application might also have to contend with systematic \emph{domain shift} where, e.g., the learner's embodiment differs systematically from that of the demonstrator, and therefore the learner's context $\omega$ cannot be treated as a sample from $p(\omega)$, we leave this challenge to prior work, and instead focus on the basic problem of imitation-from-observation. This means that the context can vary between the demonstrations and the learner, but the learner's context still comes from the same distribution. We elaborate on the practical implications of this assumption in Section~\ref{sec:experiments}, and discuss how it might be lifted in future work.

Any algorithm for imitation-from-observation must contend with two challenges: first, it must be able to determine what information from the observations to track in its own context $\omega_l$, which may differ from those of the demonstrations, and second, it must be able to determine which actions will allow it to track the demonstrated observations. Reinforcement learning (RL) offers a tool for addressing the latter problem: we can use some measure of distance to the demonstration as a reward function, and learn a policy that takes actions to minimize this distance. But which distance to use? If the observations correspond, for example, to raw image pixels, a Euclidean distance measure may not give a well-shaped objective: roughly matching pixel intensities does not necessarily correspond to a semantically meaningful execution of the task, unless the match is almost perfect. Fortunately, the solution to the first problem -- context mismatch -- naturally lends us a solution to the problem of choosing a distance metric. In order to address context mismatch, we can train a model that explicitly translates demonstrations from one context into another, by using the different demonstrations as training data. The internal representation learned by such a model provides a much more well-structured space for evaluating distances between observations, since proper context translation requires understanding the underlying factors of variation in the scene. As we empirically illustrate in our experiments, we can use squared Euclidean distances between features of the context translation model as a reward function to learn the demonstrated task, while using the model itself to translate these features from the demonstration context to the learner's context. We first describe the translation model, and then show how it can be used to create a reward function for RL.

\section{Learning to Translate Between Contexts}
\label{sec:translationmodel}

Since each demonstration $D_k$ is generated from an unknown context $\omega_k$, the learner cannot directly track these demonstrations in its own context $\omega_l$. However, since we have demonstrations from multiple unknown but different contexts, we can learn a context translation model on these demonstrations without any explicit knowledge of the context variables themselves. We only assume that the first frame $o_0^k$ of a demonstration in a particular context $\omega_k$ can be used to implicitly extract information about the context $\omega_k$. 

Our translation model is trained on pairs of demonstrations $D_i = [o_0^i, o_1^i..., o_T^i]$ and $D_j = [o_0^j, o_1^j..., o_T^j]$, where $D_i$ comes from a context $\omega_i$ (the source context) and $D_j$ comes from a context $\omega_j$ (the target context). The model must learn to output the observations in $D_j$ conditioned on $D_i$ and the first observation $o_0^j$ in the target context $\omega_j$. Thus, the model looks at a single observation from a target context, and predicts what future observations in that context will look like by translating a demonstration from a source context. Once trained, this model can be provided with any demonstration $D_k$ to translate it into the learner's context $\omega_l$ for tracking, as discussed in the next section.

\begin{figure}[!h]
  \centering
  \includegraphics[width=0.41\textwidth]{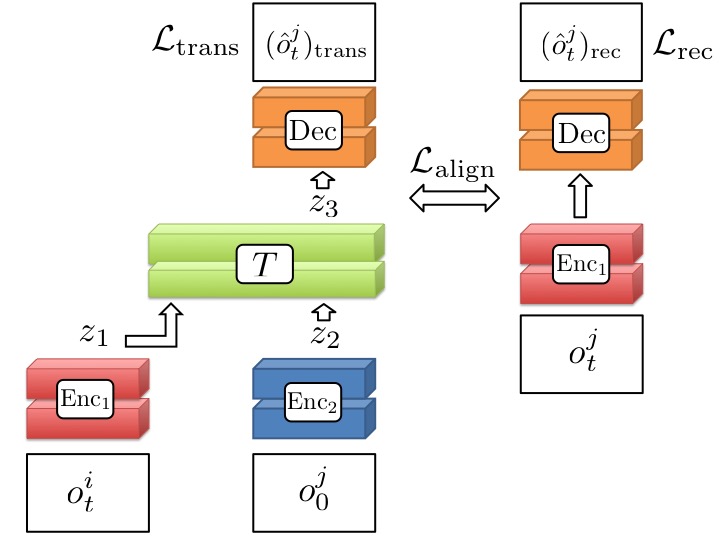} 
  {\footnotesize
  \caption{Context translation model: The source observation $o_t^i$ is translated to give the prediction of the observation in the target context $(\hat{o}_t^j)_{\text{trans}}$, given the context image $o_0^j$ from the target context. The convolutional encoders are $\text{Enc}_1$ and $\text{Enc}_2$, while the deconvolutional decoder $\text{Dec}$ decodes features back into observations. Colors indicate tied weights.
  }
  \label{fig:model}
  }
  
\end{figure}
The model (Fig~\ref{fig:model}), assumes that the demonstrations $D_i$ and $D_j$ are aligned in time, though this assumption could be relaxed in future work by using iterative time alignment~\cite{InvariantFeats}. The goal is to learn the overall translation function $M(o_t^i, o_0^j)$ such that its output $M(o_t^i, o_0^j) = (\hat{o}_t^j)_{\text{trans}}$ closely matches $o_t^j$ for all $t$ and each pair of training demonstrations $D_i$ and $D_j$. That is, the model translates observations from $D_i$ into the context $\omega_j$, conditioned on the first observation $o_0^j$ in $D_j$.

\newcommand{\vcenteredinclude}[1]{\begingroup
\setbox0=\hbox{#1}%
\parbox{\wd0}{\box0}\endgroup}

The model consists of four components: a source observation encoder $\text{Enc}_1(o_t^i)$ and a target initial observation encoder $\text{Enc}_2(o_0^j)$ that encode the observations into source and target features, referred to as $z_1$ and $z_2$, a translator $z_3 = T(z_1,z_2)$ that translates the features $z_1$ into features for the context of $z_2$, which are denoted $z_3$, and finally a target context decoder $\text{Dec}(z_3)$, which decodes these features into $\hat{o}_t^j$. We will use $F(o_t^i, o_0^j) = z_3$ to denote the feature extractor that generates the features $z_3$ from an input observation and a context image. The encoders $\text{Enc}_1$ and $\text{Enc}_2$ can have either different weights or tied weights depending on the diversity of the demonstration scenes. To deal with the complexities of pixel-level reconstruction, we include skip connections from $\text{Enc}_2$ to $\text{Dec}$. The model is supervised with a squared error loss $\mathcal{L}_{\text{trans}} = \|(\hat{o}_t^j)_{\text{trans}} - o_t^j\|^2_2$ on the output $o_t^j$ and trained end-to-end.

However, we need the features $z_3$ to carry useful information, in order to provide an informative distance metric between demonstrations for feature tracking. To ensure that the translated features $z_3$ form a representation that is internally consistent with the encoded image features $z_1$, we jointly train the translation model encoder $\text{Enc}_1$ and decoder $\text{Dec}$ as an autoencoder, with a reconstruction loss $\mathcal{L}_{\text{rec}} = \|\text{Dec}(\text{Enc}_1(o_t^j)) - o_t^j\|^2_2$. We simultaneously regularize the feature representation of this autoencoder to align it with the features $z_3$, using the loss $\mathcal{L}_{\text{align}} = \|z_3 - \text{Enc}_1(o_t^j)\|_2^2$. This forces the encoder $\text{Enc}_1$ and decoder $\text{Dec}$ to adopt a consistent feature representation, so that the observation from the target context $o_t^j$ is encoded into features that are similar to the translated features $z_3$. The training objective for the entire model is then given by the combined loss function $\mathcal{L} = \sum_{(i,j)}( \mathcal{L}_{\text{trans}} +  \lambda_1\mathcal{L}_{\text{rec}} + \lambda_2\mathcal{L}_{\text{align}})$, with  $D_i$ and $D_j$ being a pair of expert demonstrations chosen randomly from the training set, and $\lambda_1$ and $\lambda_2$ being hyperparameters. If we don't regularize the encoded features of learning trajectories and translated features of experts to lie in the same feature space, the reward function described in Section~\ref{sec:featcost} is not effective since we are tracking features which have no reason to be in the same space. Examples of translated demonstrations are shown in Section~\ref{sec:experiments} and the project website.

\section{Learning Policies via Context Translation}
\label{sec:policies}

The model described in the previous section can translate observations and features from the demonstration context into the learner's context $\omega_l$. However, in order for the learning agent to actually perform the demonstrated behavior, it must be able to acquire the actions that track the translated features. We can choose between a number of deep reinforcement learning algorithms to learn to output actions that track the translated demonstrations given the reward function we describe below.

\subsection{Reward Functions for Feature Tracking}
\label{sec:featcost}

The first component of the feature tracking reward function is a penalty for deviations from the translated features. At each time step, the translation function $F$ (which gives us $z_3$) can be used to translate each of the demonstration observations $o_t^i$ into the learner's context $\omega_l$. The reward function then corresponds to minimizing the squared Euclidean distance between the encoding of the current observation to all of these translated demonstration features, which is approximately tracking their average, resulting in
\begin{equation*}
    \hat{R}_\text{feat}(o_t^l) = -\|\text{Enc}_1(o_t^l) - \frac{1}{n} \sum_i^n F(o_t^i, o_0^l)\|_2^2,
\end{equation*}
\noindent where $\text{Enc}_1(o_t^l)$ computes the features of the learner's observation at time step $t$, given by $o_t^l$, and $F(o_t^i, o_0^j)$ computes translated features of experts.

Unfortunately, feature tracking by itself may be insufficient to successfully imitate complex behaviors. The reason for this is that the distribution of observations fed into $\text{Enc}_1$ during policy learning may not match the distribution from the demonstrations that are seen during training: although a successful policy will closely track the translated policy, a poor initial policy might produce observations that are very different. In these cases, the encoder $\text{Enc}_1$ must contend with out-of-distribution samples, which may not be encoded properly. In fact, the model will be biased toward predicting features that are closer to the expert, since it only saw expert data during training. To address this, we also introduce a weak image tracking reward. This reward directly penalizes the policy for experiencing observations that differ from the translated observations, using the full observation translation model $M$:
\begin{equation*}
  \hat{R}_\text{img}(o_t^l) = -\|o_t^l - \frac{1}{n} \sum_i^n M(o_t^i, o_0^l)\|_2^2 
\end{equation*}
The final reward is then the weighted combination $\hat{R}(o_t^l) = \hat{R}_\text{feat}(o_t^l)+w_\text{rec}\hat{R}_\text{img}(o_t^l)$, where $w_\text{rec}$ is a small constant (tuned as a hyperparameter in our implementation).

\subsection{Reinforcement Learning Algorithms for Feature Tracking}
\label{sec:RL}
With the reward described in Section ~\ref{sec:featcost}, we perform reinforcement learning in order to learn control policies in our learning environment. Our method can be used with any reinforcement learning algorithm. We use trust region policy optimization (TRPO)~\cite{TRPO} for our simulated experiments but not for real world experiments because of it's high sample complexity. For the real-world robotic experiments, we use the trajectory-centric RL method used for local policy optimization in guided policy search (GPS)~\cite{levinefinn16JMLR}, which is based on fitting locally linear dynamics and performing LQR-based updates. We compute image features $z_3$, and include these as part of the state. The cost function for GPS is then a squared Euclidean distance in state space, and we omit the image tracking cost described in Section~\ref{sec:featcost}. For simulated striking and real robot pushing, this cost function is also weighted by a quadratic ramp function weighting squared Euclidean distances at later time steps higher than initial ones.

\begin{figure}[!h]
\centering
  \includegraphics[width=0.115\textwidth]{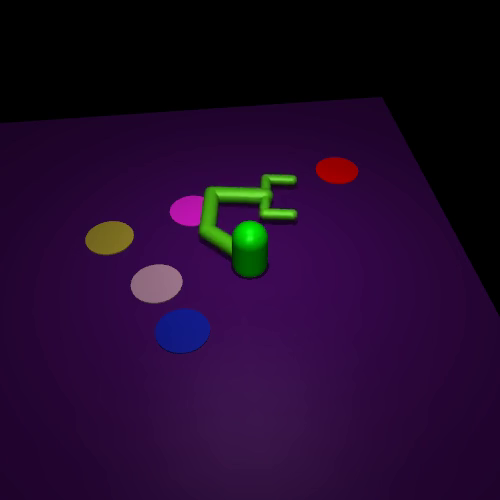} \includegraphics[width=0.115\textwidth]{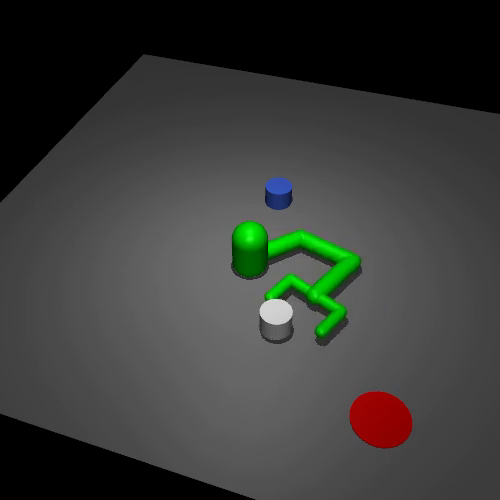} 
 \includegraphics[width=0.115\textwidth]{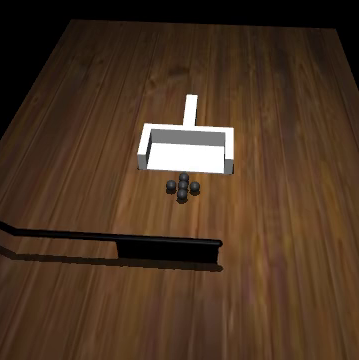} \includegraphics[width=0.115\textwidth]{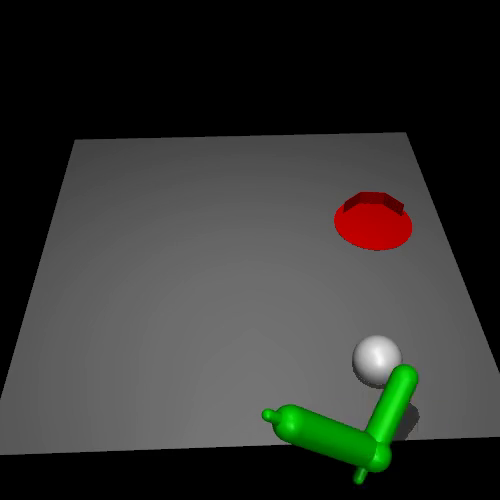}
\caption{Four simulated tasks, from left to right: reaching (goal is to reach the red circle), pushing (goal is to push the white can to the red goal), sweeping (goal is to sweep grey balls into the pan), and striking (goal is to strike the white ball to the red goal). }
\label{fig:tasks}
\end{figure}

\section{Experiments}
\label{sec:experiments}

Our experiments aim to evaluate whether our context translation model can enable imitation-from-observation, and how well representative prior methods perform on this type of imitation learning task. The specific questions that we aim to answer are: (1) Can our context translation model handle raw image observations, changes in viewpoint, and changes in the appearance and positions of objects between contexts? (2) How well do prior imitation learning methods perform in the presence of such variation, in comparison to our approach? (3) How well does our method perform on real-world images, and can it enable a real-world robotic system to learn manipulation skills? All results, including illustrative videos, video translations and further experiment details can be found on: \url{https://sites.google.com/site/imitationfromobservation/}

In order to provide detailed comparisons with alternative prior methods for imitation learning, we set up four simulated manipulation tasks using the MuJoCo simulator~\cite{mujoco}. To provide expert demonstrations, we hand-specified a reward function for each task and used a prior policy optimization algorithm~\cite{TRPO} to train an expert policy. We collected video demonstrations of rollouts from the final expert policy acting in a number of randomly generated contexts.

The tasks are illustrated in Fig.~\ref{fig:tasks}. The first task requires a robotic arm to reach varying goal positions indicated by a red disk, in the presence of variation in color and appearance.
The second task requires pushing a white cylinder onto a red coaster, both with varying position, in the presence of varied distractor objects. The third task requires the simulated robot to sweep five grey balls into a dustpan, under variation in viewpoint. The fourth task involves using a 4 DoF arm to strike a white ball toward a red target which varies in position. The project website illustrates the variability in appearance and object positioning in the tasks, and also presents example translations of individual demonstration sequences.

\subsection{Network Architecture and Training}
For encoders $\text{Enc}_1$ and $\text{Enc}_2$ in simulation we perform four $5\times5$ stride-2 convolutions with filter sizes 64, 128, 256, and 512 followed by two fully-connected layers of size 1024. We use LeakyReLU activations with leak 0.2 for all layers. The translation module $T(z_1, z_2)$ consists of one hidden layer of size 1024 with input as the concatenation of $z_1$ and $z_2$. For the decoder $\text{Dec}$ in simulation we use a fully connected layer from the input to four fractionally-strided convolutions with filter sizes 256, 128, 64, 3 and stride $\frac{1}{2}$. We have skip connections from every layer in the context encoder $\text{Enc}_2$ to its corresponding layer in the decoder $\text{Dec}$ by concatenation along the filter dimension. 
For real world images, the encoders perform 4 convolutions with filter sizes 32, 16, 16, 8 and strides 1, 2, 1, 2 respectively. All fully connected layers and feature layers are size 100 instead of 1024. The decoder uses fractionally-strided convolutions with filter sizes 16, 16, 32, 3 with strides $\frac{1}{2}$, 1, $\frac{1}{2}$, 1 respectively. For the real world model only, we apply dropout for every fully connected layer with probability 0.5, and we tie the weights of $\text{Enc}_1$ and $\text{Enc}_2$. 

We train using the ADAM optimizer with learning rate $10^{-4}$. We train using 3000 videos for reach, 4500 videos for simulated push, 894 videos for sweep, 3500 videos for strike, 135 videos for real push, 85 videos for real sweep with paper, 100 videos for real sweep with almonds, and 60 videos for ladling almonds. We downsample videos to $36\times64$ pixels for simulated sweeping and $48\times48$ for all other videos.

\begin{figure*}[!t]
  \centering
  \includegraphics[width=0.75\textwidth]{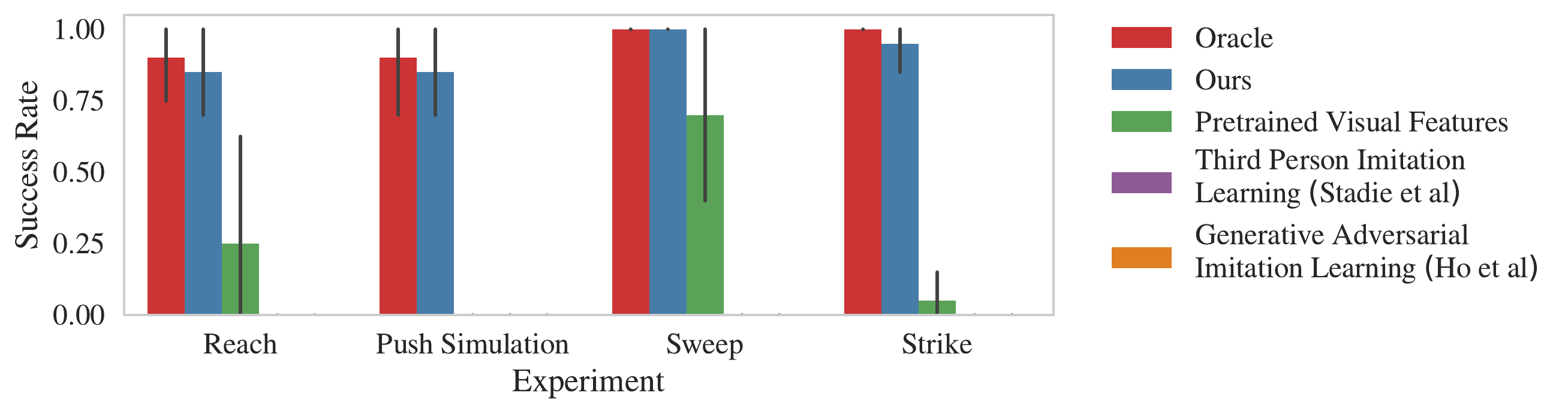}  
  \caption{Comparisons with prior methods on the reaching, pushing, sweeping, and striking tasks. The results show that our method successfully learned each task, while the prior methods struggled to perform the reaching, pushing and striking tasks, and only the pretrained visual features approach was able to make a reasonable improvement on the sweeping task. Third person imitation learning~\cite{TPIL} and generative adversarial imitation~\cite{GAIL} learning are both at 0$\%$ success rate on the graph.}
  \label{fig:allcomp}
\end{figure*}

\subsection{Comparative Evaluation of Context Translation}
\label{sec:compare}

Results for the comparative evaluation of our approach are presented in Fig~\ref{fig:allcomp}. Performance is evaluated in terms of the final distance of the target object to the goal during testing. In the reaching task, this is the distance of the robot's hand from the goal, in the pushing task, this is the distance of the cylinder from the goal, in the sweeping task, this corresponds to the mean distance of the balls from the inside of the dustpan, and in the striking task this is the final distance of the ball from the goal position. All distances are normalized by dividing by the initial distance at the start of the task, and success is measured as a thresholding of the normalized distance. We evaluate each task on 10 randomly generated environment conditions, each time performing 100 iterations of reinforcement learning with 12,500 samples per iteration.

\begin{figure*}[b!]
\centering
  \includegraphics[width=0.7\textwidth]{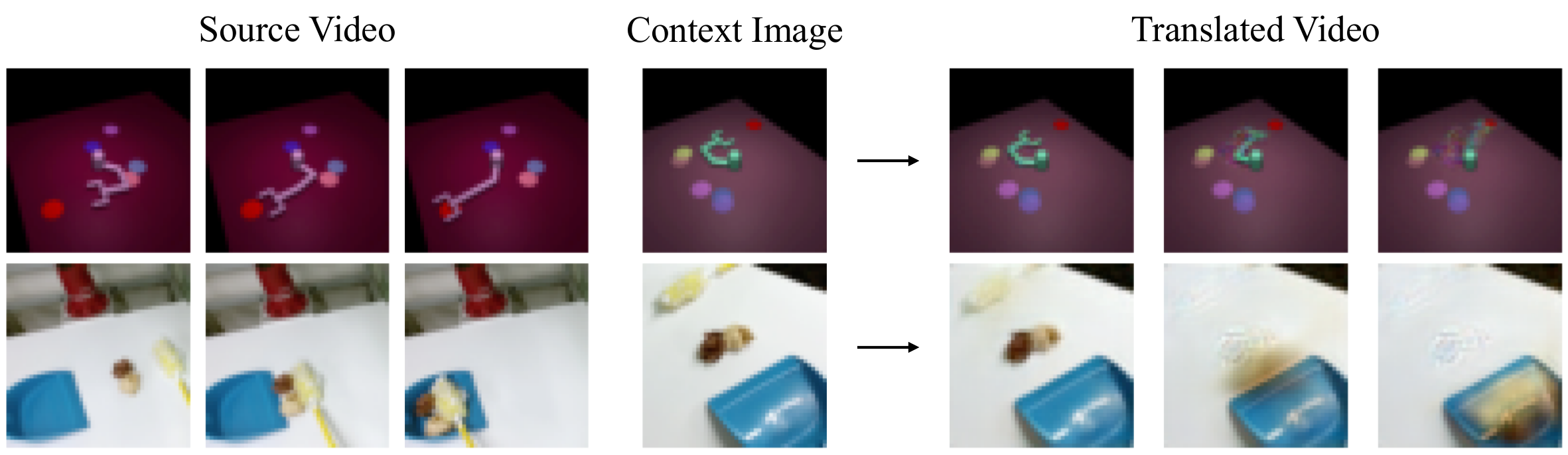} \\
  \caption{Translations from a video in the holdout set to a new context for the reaching task (top) and paper sweeping task (bottom).}
  \label{fig:realtranslations}
\end{figure*}

Our comparisons include our method with TRPO for policy learning, an oracle that trains a policy with TRPO on the ground truth reward function in simulation, which represents an upper bound on performance, and three prior imitation learning methods. The first prior method learns a reward using pre-trained visual features, similar to the work of \cite{sermanetRSS}. In this method, features from an Inception-v3 network trained on ImageNet classification~\cite{DBLP:journals/corr/SzegedyVISW15} are used to encode the goal image from the demonstration, and the reward function corresponds to the distance to these features averaged over all the training demonstrations. We experimented with several different feature layers from the Inception-v3 network and chose the one that performed best. The second prior method, third person imitation learning (TPIL), is an IRL algorithm that explicitly compensates for domain shift using an adversarial loss~\cite{TPIL}, and the third is an adversarial IRL-like algorithm called generative adversarial imitation learning (GAIL)~\cite{GAIL}, using a convolutional model to process images as suggested by~\cite{infoGAIL}. Among these, only TPIL explicitly addresses changes in domain or context.

The results, shown in Fig~\ref{fig:allcomp}, indicate that our method was able to successfully learn each of the tasks when the demonstrations were provided from random contexts. Notably, none of the prior methods were actually successful on the reaching, pushing or striking tasks, and struggled to perform the sweeping task. This indicates that imitation-from-observation in the presence of context differences is an exceedingly challenging problem.

\begin{figure}[t]
\centering
  \includegraphics[width=\linewidth]{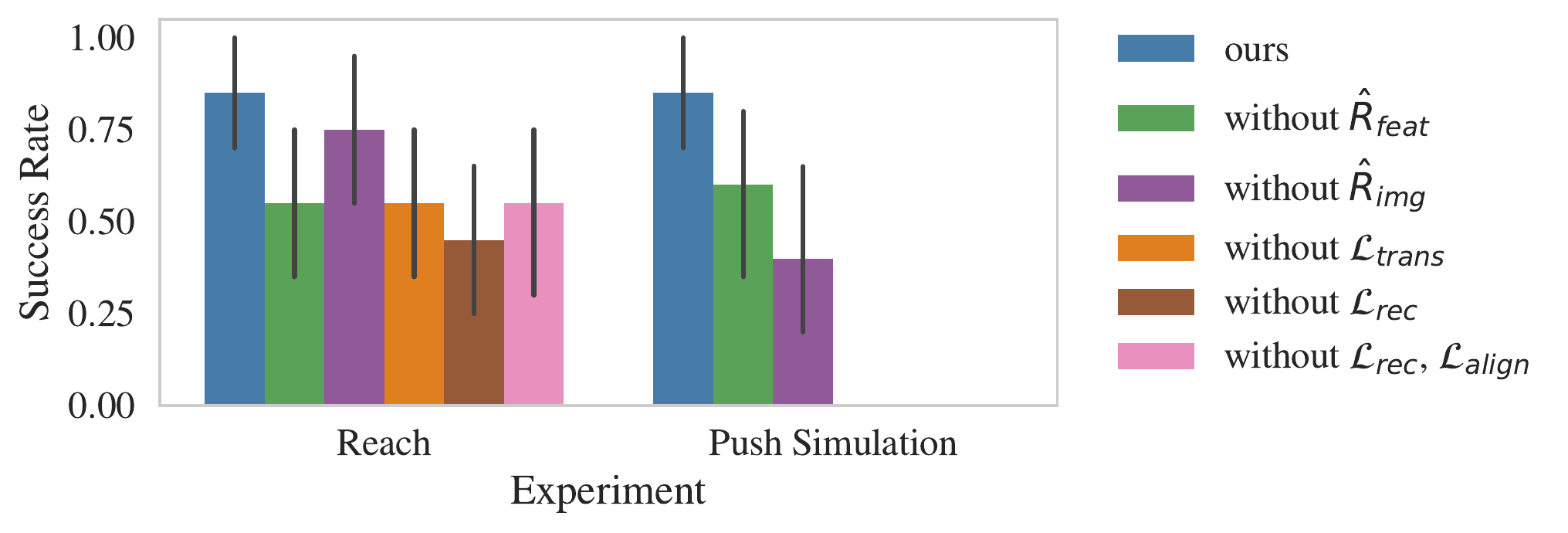} 
 
  \caption{Ablations on model losses and reward functions for the simulated reaching and pushing tasks.Our method with all components does consistently the best across tasks. Note: for the Push Simulation task, we did not perform ablations "without $\mathcal{L}_{trans}$", "without $\mathcal{L}_{rec}$", and "without $\mathcal{L}_{rec}, \mathcal{L}_{align}$"}
  \label{fig:allablations}
\end{figure}

\begin{figure*}[!t]
\centering

\includegraphics[width=0.9\textwidth]{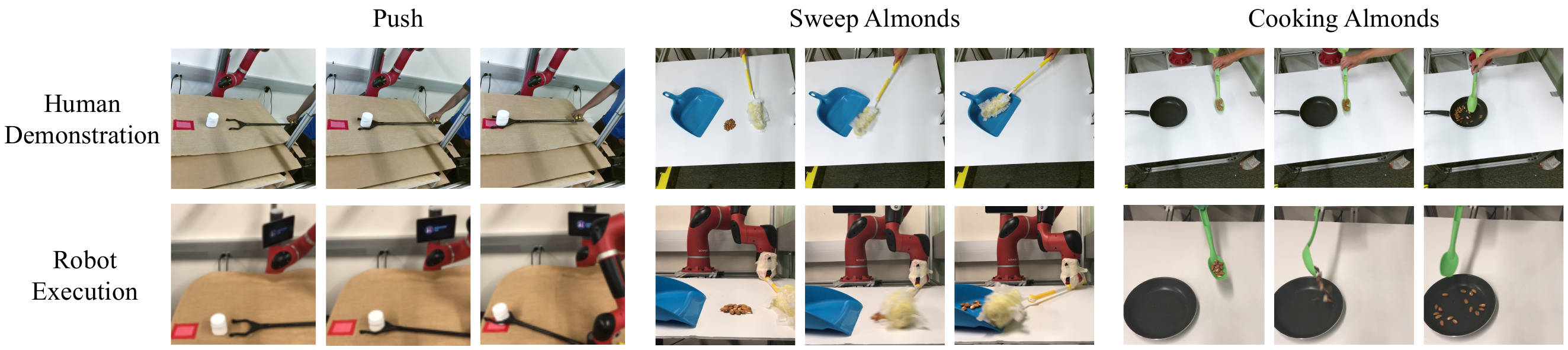}
  \caption{Top Row: Demonstrations by a human demonstrator showing the robot how to perform the pushing, sweeping and ladling almonds task in the real world. Bottom Row: Execution of the robot successfully performing the pushing, sweeping and ladling almonds tasks.}
  \label{fig:realsweepalmond}
\end{figure*}

\subsection{Ablation Study}
To evaluate the importance of different loss functions while training our translation model, and the different components for the reward function while performing imitation, we performed ablations by removing these components one by one during model training or policy learning. To understand the importance of the translation cost, we remove cost $\mathcal{L}_{\text{trans}}$, to understand whether features $z_3$ need to be properly aligned we remove model losses $\mathcal{L}_{\text{rec}}$ and $\mathcal{L}_{\text{align}}$. In Fig~\ref{fig:allablations} we see that the removal of each of these losses significantly hurts the performance of subsequent imitation. On removing the feature tracking loss $\hat{R}_\text{feat}$ or the image tracking loss $\hat{R}_\text{image}$ we see that overall performance across tasks is worse.

\subsection{Natural Images and Real-World Robotic Manipulation}
\label{sec:realexperiments}
To evaluate whether our method is able to scale to real-world images and robots, we focus on manipulation tasks involving tool use, where object positions and camera viewpoints differ between contexts. All demonstrations were provided by a human, while the learned skills were performed by a robot. Since our method assumes that the contexts of the demonstrations and the learner are sampled from the same distribution, the human and the robot both used the same tool to perform the tasks, avoiding any systematic domain shift that might result from differences in the appearance of the human or robot arm. To this end, we apply a cropping of each video around task-relevant areas of each demonstration. Investigating domain shift is left for future work, and could be done, for example, using domain adaptation~\cite{tzengdevin16WAFR}. In the present experiments, we assume that the demonstration and test contexts come from the same distribution, which is a reasonable assumption in settings such as tool use and navigation, or tasks where the focus is on the objects in the scene rather than the arm or end-effector.

\begin{figure}[h]
  \centering
  \includegraphics[width=0.5\textwidth]{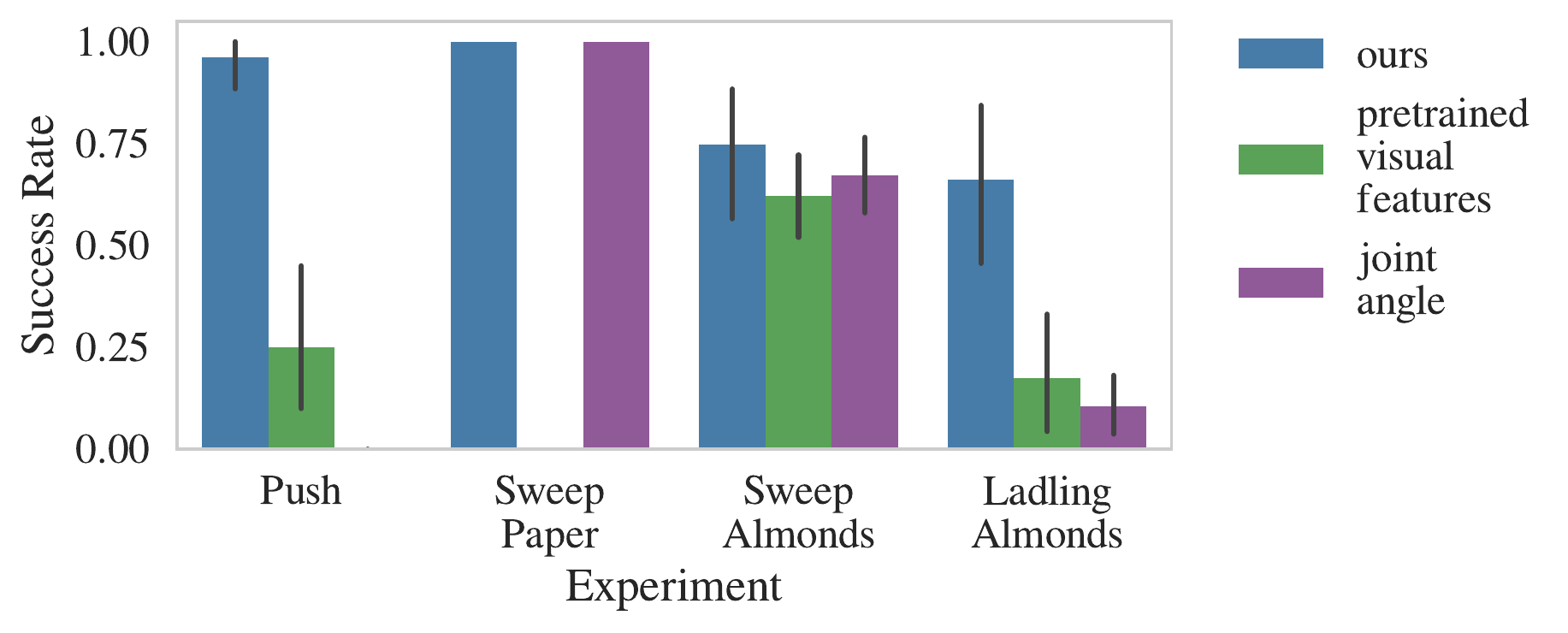} 
  \caption{Plot depicting success rate for our method versus other baselines on the real world tasks with the Sawyer robot. Success metrics differ per task as described in Section \ref{sec:realexperiments}. As is seen clearly, our method consistently performs well on all the real world tasks, and outperforms the baseline methods.
}
  \label{fig:realrobot}
  \vspace{-0.3cm}
\end{figure}

\subsubsection{Pushing}

In the first task, the goal is to push a cylinder to a marked goal position. The success metric is defined as whether the final distance between the cylinder and goal is within a predefined threshold. We evaluate our method in the setting where real-world demonstrations are provided by a human and imitation is done by a robot in the real world.

We evaluated how our method can be used to learn a pushing behavior with a real-world robotic manipulator, using a 7-DoF Sawyer robot. Since the TRPO algorithm is too data intensive to learn on real-world physical systems, we use GPS for policy learning (Section~\ref{sec:RL}). 

For comparison, we also test GPS with a reward that involves tracking pre-trained visual features from the Inception-v3 network (Section~\ref{sec:compare}), as well as a baseline reward function that attempts to reach a fixed set of joint angles, specified through kinesthetic demonstration. Note that our method itself does not use any kinesthetic demonstrations, only video demonstrations provided by the human. In order to include the high-dimensional visual features in the state for guided policy search, we apply PCA to reduce their dimensionality from 221952 to 100,
while our method uses all 100 dimensions of $z_3$ as part of the state. We found that our method could successfully learn the skill using demonstrations from different viewpoints, and outperforms the pre-trained features and kinesthetic baseline, as shown in Fig~\ref{fig:realrobot}.

\subsubsection{Sweeping}

The pushing task illustrates the basic capability of our method to learn skills involving manipulation of rigid objects. However, one major advantage of learning visual reward functions from demonstration is the ability to acquire representations that can be used to manipulate scenes that are harder to represent analytically, such as granular media. In this next experiment, we study how well our method can learn two variants of a sweeping task: in the first, the robot must sweep crumpled paper into a dustpan, and in the second it must sweep a pile of almonds. We used almonds in place of dirt or fluids to avoid damaging the robot. Quantitative results are summarized in Fig~\ref{fig:realrobot}.

On the easier crumpled paper task, both our method and the kinesthetic teaching approach works well, but the reward that uses pre-trained visual features is insufficient to accomplish the task. On the almond sweeping task (Fig~\ref{fig:realsweepalmond}), our method achieves a higher success rate than the alternative approaches. The success metric is defined as the average percentage of almonds or paper pieces that end up inside the dustpan.

\subsubsection{Ladling Almonds}

Our last task combines granular media (almonds) and a more dynamic behavior. In this task, the robot must ladle almonds into a cooking pan (Fig ~\ref{fig:realsweepalmond}). This requires keeping the ladle upright until over the pan, and then dumping them into the pan by turning the wrist. The success metric is the average fraction of almonds that were ladled into the pan. Learning from only raw videos of the task being performed by a human in different contexts, our method achieved a success rate of 66$\%$ , while the alternative approaches generally could not perform this task. An insight into why the joint angles approach wouldn't work on this task is that the spoon has to remain upright until just the right position over the pan after which it should rotate and pour into the pan. The joint angle baseline can simply interpolate between the final turned spoon position and the initial position and pour the almonds in the wrong location. Quantitative results and comparisons are summarized in Fig~\ref{fig:realrobot}.

\section{Discussion and Future Work}

We investigated how imitation-from-observation can be performed by learning to translate demonstration observation sequences between different contexts, such as differences in viewpoint. After translating observations into a target context, we can track these observations with RL, allowing the learner to reproduce the observed behavior. The translation model is trained by translating between the different contexts observed in the training set, and generalizes to the unseen context of the learner. Our experiments show that our method can be used to perform a variety of manipulation skills, and can be used for real-world robotic control on a diverse range of tasks patterned after common household chores.

Although our method performs well on real-world tasks and several tasks in simulation, it has a number of limitations. First, it requires a substantial number of demonstrations to learn the translation model. Training an end-to-end model from scratch for each task may be inefficient in practice, and combining our method with higher level representations proposed in prior work would likely lead to more efficient training~\cite{sermanetRSS,tcn}. Second, we require observations of demonstrations from multiple contexts in order to learn to translate between them. In practice, the number of available contexts may be scarce. Future work would explore how multiple tasks can be combined into a single model, where different tasks might come from different contexts. Finally, it would be exciting to explore explicit handling of domain shift in future work, so as to handle large differences in embodiment and learn skills directly from videos of human demonstrators obtained, for example, from the Internet.

\bibliographystyle{IEEEtran}
\bibliography{references}
\end{document}

\appendix
\section*{Appendix A. Network Architecture and Training}
For encoders $\text{Enc}_1$ and $\text{Enc}_2$ in simulation we use stride-2 convolutions with a $5\times5$ kernel. We perform 4 convolutions with filter sizes 64, 128, 256, and 512 followed by two fully-connected layers of size 1024. We use LeakyReLU activations with leak 0.2 for all layers. The translation module $T(z_1, z_2)$ consists of one hidden layer of size 1024 with input as the concatenation of $z_1$ and $z_2$ and output of size 1024. For the decoder $\text{Dec}$ in simulation we have a fully connected layer from the input to four fractionally-strided convolutions with filter sizes 256, 128, 64, 3 and stride $\frac{1}{2}$. We have skip connections from every layer in the context encoder $\text{Enc}_2$ to its corresponding layer in the decoder $\text{Dec}$ by concatenation along the filter dimension. 

For real world images, the encoders perform 4 convolutions with filter sizes 32, 16, 16, 8 and strides 1, 2, 1, 2 respectively. All fully connected layers and feature layers are size 100 instead of 1024. The decoder uses fractionally-strided convolutions with filter sizes 16, 16, 32, 3 with strides $\frac{1}{2}$, 1, $\frac{1}{2}$, 1 respectively. For the real world model only, we apply dropout for every fully connected layer with keep probability 0.5, and we tie the weights of $\text{Enc}_1$ and $\text{Enc}_2$. 

We train using the ADAM optimizer with learning rate $10^{-4}$. We train using 3000 videos for reach, 4500 videos for simulated push, 894 videos for sweep, 180 videos for simulated push with real videos, and 135 videos for real push with real videos.

\section*{Appendix B. Ablation Study}
To evaluate that the different loss functions while training our translation model, and the different components for the reward function while performing imitation, we performed ablations by removing these components one by one during model training or policy learning. To understand the importance of the translation cost, we remove cost $\mathcal{L}_{\text{trans}}$, to understand whether features $z_3$ need to be properly aligned we remove model losses $\mathcal{L}_{\text{rec}}$ and $\mathcal{L}_{\text{align}}$. We see that the removal of each of these losses significantly hurts the performance of subsequent imitation. On removing the feature tracking loss $\hat{R}_\text{feat}$ or the image tracking loss $\hat{R}_\text{image}$ we see that overall performance across tasks is worse.

\begin{figure}[h!]
\centering
  \includegraphics[width=0.95\textwidth]{plot/barablation.png} 
 
  \caption{Ablations on model losses and reward functions for the reaching, pushing and pushing with real world demonstrations tasks. Across tasks, all components of the model are necessary for success.}
  \label{fig:allablations}
\end{figure}

\newpage
\section*{Appendix C. Sample Videos}

\subsection*{Reach Simulation}
\begin{figure}[h!]
  \centering
  \includegraphics[width=0.9\textwidth]{diagrams/reach/reach6.jpg}
  \includegraphics[width=0.9\textwidth]{diagrams/reach/reach8.jpg}
  \caption{Example expert training demonstrations from different viewpoints with variations in color, distractor objects, and goal position.}
\end{figure}

\begin{figure}[h!]
  \small{Source Video} \hspace*{0.73cm}
  \vcenteredinclude{
  \includegraphics[width=0.8\textwidth]{diagrams/reach/reachsrc16.jpg}}\\
  \small{Target Context $o_0$} \hspace*{0.45cm}
  \vcenteredinclude{\includegraphics[width=0.145\textwidth]{diagrams/reach/reachctx16.png}}\\
  \small{Translated Video} \hspace*{0.3cm}
  \vcenteredinclude{\includegraphics[width=0.8\textwidth]{diagrams/reach/reachtrans16.jpg}}\\
  
  \small{Source Video} \hspace*{0.73cm}
  \vcenteredinclude{
  \includegraphics[width=0.8\textwidth]{diagrams/reach/reachsrc21.jpg}}\\
  \small{Target Context $o_0$} \hspace*{0.45cm}
  \vcenteredinclude{\includegraphics[width=0.145\textwidth]{diagrams/reach/reachctx21.png}}\\
  \small{Translated Video} \hspace*{0.3cm}
  \vcenteredinclude{\includegraphics[width=0.8\textwidth]{diagrams/reach/reachtrans21.jpg}}\\
  \caption{Example illustrations of demonstrations for a reaching task (top) being performed in a new context (middle), with the translated observation sequences (bottom).}
\end{figure}
\newpage

\subsection*{Push Simulation}
\begin{figure}[h!]
  \centering
  \includegraphics[width=0.9\textwidth]{diagrams/push/push0.jpg}
  \includegraphics[width=0.9\textwidth]{diagrams/push/push6.jpg}
  \caption{Example expert training demonstrations from different viewpoints with variations in distractor objects, start and goal position.}
\end{figure}

\begin{figure}[h!]
  \small{Source Video} \hspace*{0.73cm}
  \vcenteredinclude{
  \includegraphics[width=0.8\textwidth]{diagrams/push/push2src.jpg}}\\
  \small{Target Context $o_0$} \hspace*{0.45cm}
  \vcenteredinclude{\includegraphics[width=0.145\textwidth]{diagrams/push/push2ctx.png}}\\
  \small{Translated Video} \hspace*{0.3cm}
  \vcenteredinclude{\includegraphics[width=0.8\textwidth]{diagrams/push/push2trans.jpg}}\\
  
  \small{Source Video} \hspace*{0.73cm}
  \vcenteredinclude{
  \includegraphics[width=0.8\textwidth]{diagrams/push/push6src.jpg}}\\
  \small{Target Context $o_0$} \hspace*{0.45cm}
  \vcenteredinclude{\includegraphics[width=0.145\textwidth]{diagrams/push/push6ctx.png}}\\
  \small{Translated Video} \hspace*{0.3cm}
  \vcenteredinclude{\includegraphics[width=0.8\textwidth]{diagrams/push/push6trans.jpg}}\\
  \caption{Example illustrations of demonstrations for a pushing task (top) being performed in a new context (middle), with the translated observation sequences (bottom).}
\end{figure}

\newpage
\subsection*{Sweep Simulation}
\begin{figure}[h!]
  \centering
  \includegraphics[width=0.9\textwidth]{diagrams/sweep/sweep5.jpg}
  \includegraphics[width=0.9\textwidth]{diagrams/sweep/sweep7.jpg}
  \caption{Example expert training demonstrations from different viewpoints.}
\end{figure}

\begin{figure}[h]
  \small{Source Video} \hspace*{0.73cm}
  \vcenteredinclude{
  \includegraphics[width=0.8\textwidth]{diagrams/sweep/sweep25src.jpg}}\\
  \small{Target Context $o_0$} \hspace*{0.45cm}
  \vcenteredinclude{\includegraphics[width=0.145\textwidth]{diagrams/sweep/sweep25ctx.png}}\\
  \small{Translated Video} \hspace*{0.3cm}
  \vcenteredinclude{\includegraphics[width=0.8\textwidth]{diagrams/sweep/sweep25trans.jpg}}\\
  
  \small{Source Video} \hspace*{0.73cm}
  \vcenteredinclude{
  \includegraphics[width=0.8\textwidth]{diagrams/sweep/sweep23src.jpg}}\\
  \small{Target Context $o_0$} \hspace*{0.45cm}
  \vcenteredinclude{\includegraphics[width=0.145\textwidth]{diagrams/sweep/sweep23ctx.png}}\\
  \small{Translated Video} \hspace*{0.3cm}
  \vcenteredinclude{\includegraphics[width=0.8\textwidth]{diagrams/sweep/sweep23trans.jpg}}\\
  \caption{Example illustrations of demonstrations for a sweeping task (top) being performed in a new context (middle), with the translated observation sequences (bottom).}
\end{figure}

\newpage
\subsection*{Striking Simulation}
\begin{figure}[h!]
  \centering
  \includegraphics[width=0.8\textwidth]{diagrams/strike/strike1.jpg}
  \includegraphics[width=0.8\textwidth]{diagrams/strike/strike2.jpg}
  \caption{Example expert training demonstrations from different viewpoints.}
\end{figure}

\begin{figure}[h]
  \small{Source Video} \hspace*{0.73cm}
  \vcenteredinclude{
  \includegraphics[width=0.8\textwidth]{diagrams/strike/strikesrc1.jpg}}\\
  \small{Target Context $o_0$} \hspace*{0.45cm}
  \vcenteredinclude{\includegraphics[width=0.145\textwidth]{diagrams/strike/strikectx1.png}}\\
  \small{Translated Video} \hspace*{0.3cm}
  \vcenteredinclude{\includegraphics[width=0.8\textwidth]{diagrams/strike/striketrans1.jpg}}\\
  
  \small{Source Video} \hspace*{0.73cm}
  \vcenteredinclude{
  \includegraphics[width=0.8\textwidth]{diagrams/strike/strikesrc2.jpg}}\\
  \small{Target Context $o_0$} \hspace*{0.45cm}
  \vcenteredinclude{\includegraphics[width=0.145\textwidth]{diagrams/strike/strikectx2.png}}\\
  \small{Translated Video} \hspace*{0.3cm}
  \vcenteredinclude{\includegraphics[width=0.8\textwidth]{diagrams/strike/striketrans2.jpg}}\\
  \caption{Example illustrations of demonstrations for a striking task (top) being performed in a new context (middle), with the translated observation sequences (bottom).}
\end{figure}